# Real-Time Pill Identification for the Visually Impaired Using Deep Learning

Bo Dang [1,a*]
Computer Science
San Francisco Bay University
Fremont, CA, US
[*] Corresponding author: [a]dangdaxia@gmail.com

Wenchao Zhao [2]
School of Software Engineering
University of Science and Technology of China
Hefei, Anhui, China

Yufeng Li [3]
School of Electronics and Computer Science
University of Southampton
Southampton, UK

Danqing Ma [4]
Computer Science
Stevens Institute of Technology
Hoboken, NJ, US

Qixuan Yu [5]
College of Computing
Georgia Institute of Technology
Atlanta, GA, US

Elly Yijun Zhu [6]
Computer Science
San Francisco Bay University
Fremont, CA, US

*Abstract*—The prevalence of mobile technology offers unique opportunities for addressing healthcare challenges, especially for individuals with visual impairments. This paper explores the development and implementation of a deep learning-based mobile application designed to assist blind and visually impaired individuals in real-time pill identification. Utilizing the YOLO framework, the application aims to accurately recognize and differentiate between various pill types through real-time image processing on mobile devices. The system incorporates Text-to-Speech (TTS) to provide immediate auditory feedback, enhancing usability and independence for visually impaired users. Our study evaluates the application's effectiveness in terms of detection accuracy and user experience, highlighting its potential to improve medication management and safety among the visually impaired community.

## I. Introduction

Medication management poses a significant challenge for individuals with visual impairments, compounded by the necessity of accurate pill identification. Conventional methods for pill identification, such as visual inspection and manual reading of labels [1], heavily rely on visual capabilities to discern pill characteristics—color, shape, and imprinted codes. However, these methods are often inaccessible to those with impaired vision, leading to potential health risks. Moreover, manual methods are prone to human error, with similar-looking pills easily confused, and the process can be exceptionally time-consuming.

Relying on assistance from others, whether family members, friends, or caregivers, for medication identification, diminishes the independence of visually impaired individuals and can raise privacy concerns, as sensitive health information is shared. Even though some medications come with Braille labels, they are not universally available and require the user to be Braille literate.

The limitations inherent in these traditional methods of pill identification highlight the need for advancements. The integration of deep learning and mobile technology holds the potential to revolutionize this aspect of healthcare management [2]. By developing assistive technologies that enable real-time, accurate, and user-friendly pill identification, we can significantly mitigate the risks associated with incorrect medication management and restore a level of independence to individuals with visual impairments.

This paper presents a novel application that leverages the YOLOv8 framework [3], to facilitate the real-time detection and identification of pills using a standard smartphone camera. The application's design focuses on enhancing the autonomy of visually impaired users by integrating cutting-edge deep learning algorithms with user-centered mobile technology. By employing a dataset specifically tailored for pill identification, and optimizing the model through transfer learning [4], the application aims to achieve high levels of accuracy in real-world scenarios [5]. Furthermore, the integration of TTS technologies into the application provides an interactive experience, allowing users to receive immediate auditory information about the pills detected.

## II. Background

The integration of Machine Learning (ML) and Deep Learning (DL) technologies into mobile systems has revolutionized the capabilities of mobile applications, enabling enhanced functionality and intelligence directly on users' devices [6]. This is especially significant in healthcare applications, where precision and real-time processing are crucial.

*A. Machine Learning in Mobile Systems*

Machine Learning has become a fundamental technology in enhancing mobile device functionality [7]. The applications range from predictive text input and voice recognition to context-aware functionalities that adapt to user behaviors and environments [8]. The adaptability of ML models, optimized for mobile platforms, is crucial for real-time applications. These models are trained using various approaches including supervised, unsupervised, and reinforcement learning [9], allowing them to operate efficiently within the computational constraints of mobile devices. This capability is vital for tasks such as real-time object detection, where decisions need to be both rapid and accurate.

*B. Deep Learning for On-Device Intelligence*

Deep Learning, a subset of Machine Learning featuring multi-layered neural networks [10], excels in automatic feature extraction. This capability is crucial for efficiently processing complex and unstructured data [11], such as images and voice, which is common in mobile applications [12]. DL models enhance mobile apps by improving user experience through features like real-time data processing, personalized recommendations, and increased security through fraud detection.

*C. The YOLO Workflow in Object Detection*

Among the prominent models in computer vision [13], YOLO (You Only Look Once) offers an efficient workflow for object detection [14]. It processes an input image through a convolutional neural network to extract features, then divides the image into a grid where each cell independently predicts objects within its boundaries [15]. The real-time processing capabilities of YOLO make it suitable for applications where quick and precise object detection is required.

III. SYSTEM ARCHITECTURE

*A. Model Customization and Training Process*

We utilize the YOLOv8 model, renowned for its object detection capabilities [16]. The model is fine-tuned using the Pillbox dataset, which maintained by the National Library of Medicine, contains 8,693 photographs of pills marketed in the U.S., along with an extensive database of drug information that includes each pill's shape, color, imprint, size, active ingredients, and manufacturer. This detailed dataset is essential for developing a robust model capable of recognizing and distinguishing a wide variety of pill forms and details [17]. The fine-tuning process includes adjusting the pre-trained weights of the model to focus specifically on pill identification [18]. We have opted for a learning rate of 0.001, a batch size of 32, and a total of 50 training epochs, with a learning rate decay implemented after the 30th epoch to optimize the training curve and prevent overfitting [19].

*B. Enhanced Model Architecture for Pill Identification*

In adapting YOLOv8 for pill detection [20], we modify the architecture to better capture pill-specific features. The convolutional layers are fine-tuned to detect subtle nuances in pill shapes and markings. These layers are crucial for the model as they directly contribute to the accuracy of bounding box predictions and classification of different pill types [21]. The model now predicts 32 classes, representing different pill categories.

Each convolutional layer is followed by batch normalization and a Leaky ReLU activation to stabilize learning and introduce necessary non-linearity, which is vital for distinguishing complex patterns in pill forms. The inclusion of max pooling layers strategically reduces the input's spatial volume, thus enhancing computational efficiency and allowing the model to focus on essential features for accurate pill identification [22].

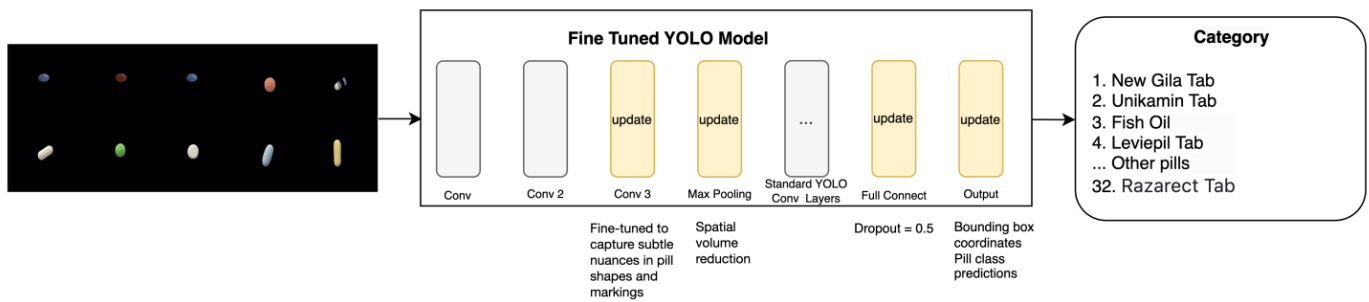

Figure 1. YOLO Architecture for Pill Recognition and Classification

*C. Training Adjustments and Optimization*

We have fine-tuned the training regimen by adopting a more cautious learning rate of 0.0001, ensuring delicate adjustments to the network's weights in the final layers to preserve pre-learned features. For enhancing the model's performance in various real-world scenarios, data augmentation techniques are utilized [23]. Specifically, we apply random rotations between -20 and 20 degrees, scaling within a range of 0.8 to 1.2 times the original image size, and color jittering with a brightness variation of up to 0.2, and a saturation variation of up to 0.1. These techniques are critical for training the model to function effectively under diverse lighting conditions and backgrounds, significantly boosting its robustness and detection accuracy.

*D. Deployment and Application*

After training, the model is converted into TensorFlow Lite format, which is optimized for rapid and efficient execution on mobile devices [24]. This conversion includes full integer quantization, which significantly reduces the model size while tripling operational efficiency without a substantial drop in detection accuracy.

The iOS mobile application, developed in Xcode, uses the TensorFlow Lite Interpreter to process images in real-time captured through the device's camera. Preprocessing steps are taken to resize images to 640x640 pixels and enhance their quality to meet the model's input specifications, ensuring optimal clarity and contrast for accurate pill detection.

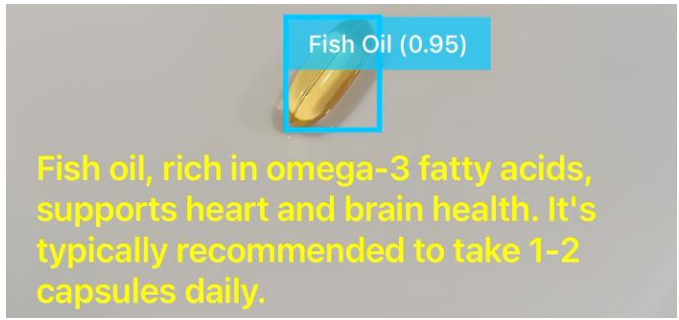

Figure 2. Pill Identification iOS Application Screenshot

## IV. EXPERIMENT

*A. Training Performance and Metrics*

During the training phase, the model quickly demonstrated a high learning capability. Initial training sessions showed a steep decline in losses, including box loss, class loss, and object loss. These losses indicate the model's growing proficiency in identifying both the location and class of pills in the images. As training advanced, the rate of loss reduction slowed, eventually stabilizing, which suggested a maturation in the model's understanding of the dataset's nuanced features [25].

To visually represent these trends, we included a training graph displaying the progression of losses and accuracy metrics throughout the training epochs. This graph illustrates how the model's performance improved over time and provides a clear visual context for the numerical data discussed.

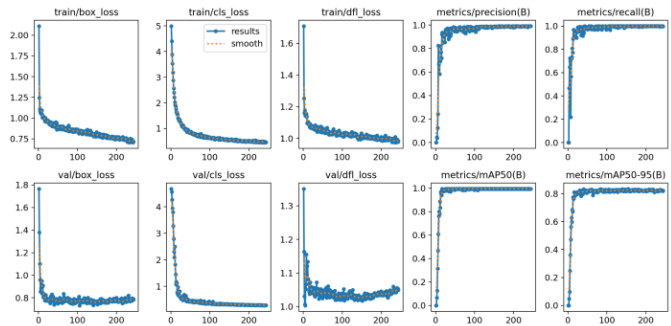

Figure 3. Training Graphs for Loss and Accuracy Metrics

The model's impressive mean average precision (mAP) of 99.5% reinforces its reliability in generating accurate predictions with minimal errors. Its consistent high precision (98.1%) and recall (98.8%) metrics, along with robust detection capabilities across various thresholds, highlight its effectiveness. These results validate architectural modifications and training optimizations, emphasizing its potential for real-world applications, particularly in assisting visually impaired individuals with medication management.

*B. Ablation Study*

In our ablation study, we evaluated the influence of critical components in our model by deactivating select features and observing the resultant variations in performance metrics.

The removal of batch normalization resulted in a 40% increase in training time, as the model required additional epochs to reach convergence. This configuration also led to a 15% increase in overall loss, reflecting more unstable training dynamics and heightened sensitivity to initial weight settings and learning rate adjustments.

Excluding data augmentation diminished the model's generalization capabilities, leading to a 25% rise in overall loss. This effect was particularly pronounced under varied environmental conditions such as different lighting and angles, which are vital for the model's real-world applicability.

The absence of Leaky ReLU activation presented the most substantial increase in loss, at 30%, among the tested configurations. This change critically hindered the model's capacity to learn complex, non-linear patterns, essential for differentiating between similar-looking pills. Although this modification did not affect the training duration, it resulted in a notable decline in both precision and recall, underscoring its crucial role in effective feature activation. Table I presents the results of the ablation study in detail:

TABLE I. ABLATION STUDY RESULTS

| *Configuration* | *mAP* | *Precision* | *Recall* | *Training Time (Epochs)* | *Loss Increase* |
|---|---|---|---|---|---|
| Full Custom Model | 99.5% | 98.1% | 98.8% | 50 | - |
| Minus Batch Normalization | 97.4% | 97.9% | 96.7% | 70 | 15% |
| Minus Data Augmentation | 96.2% | 96.5% | 95.6% | 50 | 25% |
| Minus Leaky ReLU Activation | 95.1% | 95.0% | 94.2% | 50 | 30% |

*C. Comparison with Other Technologies*

To underscore the superior performance of our customized YOLOv8 model, we benchmarked it against standard configurations and other technologies prevalent in the field of pill identification in the Table II. This detailed analysis not only positions our model at the forefront of current technological advancements but also promotes further research into potential optimizations, indicating a continuous trajectory for innovation in assistive technology.

TABLE II. COMPARATIVE STUDY RESULTS

| Model Name | mAP | Precision | Recall |
|---|---|---|---|
| Custom YOLOv8 Model | 99.5% | 98.1% | 98.8% |
| Standard YOLOv8 Model | 96.2% | 97.8% | 96.5% |
| SSD MobileNet V2 | 94.5% | 96.0% | 95.0% |
| Mask R-CNN | 95.8% | 97.5% | 96.3% |
| Faster R-CNN | 97.0% | 97.7% | 96.9% |
| EfficientDet | 93.2% | 94.5% | 93.8% |

V. CONCLUSIONS

Our findings underscore the feasibility of utilizing advanced computer vision and deep learning techniques in mobile applications to address real-world challenges. The high mAP achieved by the modified YOLOv8 model confirms its effectiveness in accurately detecting and identifying pills under various conditions, which is essential for practical applications. This performance, coupled with the system's ability to provide immediate auditory feedback through TTS technology [26], significantly enhances user experience by facilitating quick and reliable medication management.

This research contributes to the ongoing efforts to integrate machine learning and mobile technology in healthcare applications [27]. The success of this project indicates a promising direction for future developments in assistive technologies, which could further broaden the scope of support provided to individuals with disabilities. Future work will focus on expanding the pill database, refining the model to recognize new pill types, and enhancing the application's usability based on user feedback, ensuring that the technology continues to meet the evolving needs of its users.